\documentclass[11pt]{article}

\usepackage[preprint]{acl}
\usepackage{times,latexsym}
\usepackage{url}
\usepackage{amsmath}
\usepackage{amsfonts}
\usepackage[T1]{fontenc}
\usepackage{array} 
\usepackage{authblk}
\usepackage{arydshln} 
\usepackage{pifont}
\usepackage[dvipsnames]{xcolor}
\usepackage{bbding}
\usepackage[whole]{bxcjkjatype}
\usepackage{graphics}
\usepackage{graphicx}
\usepackage{multirow}
\usepackage{CJKutf8}
\usepackage{booktabs}
\usepackage{amssymb}
\usepackage{listings} 

\newcommand{\argmin}{\mathop{\rm arg~min}\limits}
\newcommand{\code}[1]{\texttt{#1}}
\usepackage[T1]{fontenc}

\usepackage[utf8]{inputenc}

\usepackage{microtype}

\usepackage{inconsolata}

\usepackage{graphicx}

\title{Routing by Analogy: kNN-Augmented Expert Assignment for Mixture-of-Experts}

\author{Boxuan Lyu$^{\text{1,2}}$, Soichiro Murakami$^{\text{2}}$, Hidetaka Kamigaito$^{\text{2,3}}$, and Peinan Zhang$^{\text{2}}$ \\
  $^{\text{1}}$Institute of Science Tokyo, $^{\text{2}}$CyberAgent, \\$^{\text{3}}$Nara Institute of Science and Technology\\
  \texttt{\url{lyu@lr.first.iir.isct.ac.jp}}\\ \texttt{\url{{murakami_soichiro,zhang_peinan}@cyberagent.co.jp}}\\ \texttt{\url{kamigaito.h@is.naist.jp}}}

\begin{document}
\maketitle

\begin{abstract}
Mixture-of-Experts (MoE) architectures scale large language models efficiently by employing a parametric ``router'' to dispatch tokens to a sparse subset of experts. 
Typically, this router is trained once and then frozen, rendering routing decisions brittle under distribution shifts. 
We address this limitation by introducing \textbf{kNN-MoE}, a retrieval-augmented routing framework that reuses locally optimal expert assignments from a memory of similar past cases. 
This memory is constructed offline by directly optimizing token-wise routing logits to maximize the likelihood on a reference set. 
Crucially, we use the average similarity of retrieved neighbors as a confidence-driven mixing coefficient, thus allowing the method to fall back to the frozen router when no relevant cases are found. 
Experiments show that kNN-MoE outperforms the zero-shot baseline and is competitive with computationally intensive supervised fine-tuning.
\end{abstract}

\section{Introduction}
\begin{figure}[t]
    \centering
    \includegraphics[width=1\linewidth]{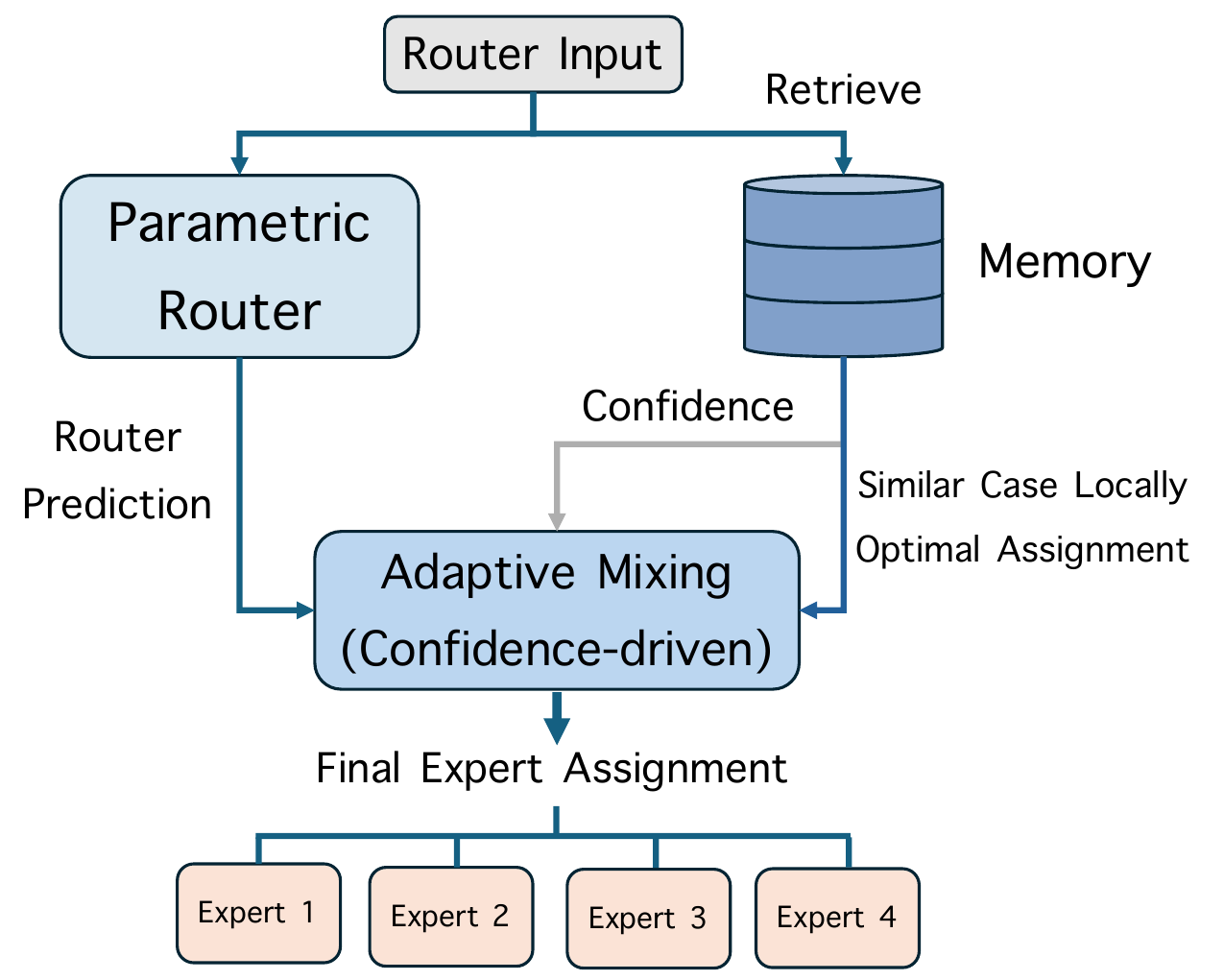}
    \caption{Schematic illustration of the \textbf{kNN-MoE} inference mechanism. 
    Given a router input, the model simultaneously obtains a prediction from the frozen ``Parametric Router'' and retrieves similar past cases from the ``Memory''. The ``Adaptive Mixing'' module fuses these two signals based on a confidence score (derived from the average similarity of retrieved neighbors) to produce the final expert assignment.}
    \label{fig:overview}
\end{figure}

The scaling of Large Language Models (LLMs) has been significantly advanced by combining Transformers \citep{transformer} with Mixture-of-Experts (MoE) architectures \citep{moe_1, moe_2, moe_3, switch_transformer}, which increase model capacity via sparse activation without a proportional increase in computational cost \citep{switch_transformer}.
A routing network (or ``router'') dynamically activates a small subset of experts for each input token \citep{switch_transformer, mixtral}. 
However, the efficacy of an MoE model hinges critically on the quality of its routing decisions.

Standard routers are lightweight, parametric classifiers trained to predict expert assignments based on hidden states.
Once trained, the router implements a fixed policy. 
While effective on in-distribution data, it lacks the flexibility to adjust routing decisions during inference. 
When facing inputs that deviate from the training distribution---often manifested as high perplexity---the router must extrapolate from fixed priors, potentially selecting suboptimal experts \citep{moe_opt1, moe_opt2, moe_opt3}. 
This lack of flexibility restricts the potential of MoE models, particularly on specialized or out-of-distribution tasks.

In this paper, we propose \textbf{kNN-MoE}, which refines routing decisions by retrieving the nearest neighbors \citep{knn} of past router inputs and reusing their offline-computed locally optimal expert assignments (Figure~\ref{fig:overview}).
Concretely, we build a memory that stores router inputs (keys) paired with locally optimal expert assignments (values) obtained by maximizing the likelihood on a reference set.
At test time, kNN-MoE retrieves similar keys from the memory whose expert assignments led to higher past case likelihoods and then interpolates the router prediction with the retrieved expert assignment, using similarity as a confidence score.
Intuitively, when retrieved neighbors are highly similar, the method trusts the retrieved routing; otherwise, it falls back to the original parametric router, avoiding noise from irrelevant retrievals.

We evaluated kNN-MoE on three MoE models---OLMoE \citep{olmoe}, GPT-OSS \citep{gptoss}, and Qwen3 \citep{qwen3}---across benchmarks including GPQA \citep{gpqa}, SuperGPQA \citep{supergpqa}, MMLU \citep{mmlu}, medical tasks (USMLE \citep{usmle} and MedMCQA \citep{medmcqa}) and a code task (MBPP \citep{mbpp}). 
Empirical results show that kNN-MoE outperforms the zero-shot baseline and achieves performance competitive with computationally intensive supervised fine-tuning, all without modifying the model parameters.
Ultimately, kNN-MoE bridges the gap between static parametric routing and dynamic inference needs, offering a scalable pathway to enhance MoE adaptability without the computational burden of fine-tuning.

\section{Related Work}
\label{sec:rw}

\subsection{Expert Assignment Optimization for MoE}
Prior work has demonstrated that optimizing expert assignments in MoE can enhance model performance. 
\citet{moe_opt3} proposed C3PO, a test-time adaptation method that fine-tunes routing logits using examples correctly answered by the model. 
However, this approach incurs a computational overhead approximately five times greater than that of the baseline \citep{moe_opt3}.
\citet{moe_opt2} proposed fine-tuning the router via an auxiliary loss; unlike our non-parametric approach, which maintains a frozen model, their method requires permanently updating the router parameters. 
\citet{moe_opt1} introduced a test-time adaptation framework that dynamically optimizes expert assignments by minimizing loss on the input context. 
In contrast, kNN-MoE eliminates this online optimization overhead by directly retrieving cached locally optimal assignments from a memory store.

Notably, ERC loss \citep{cnm2}, has incorporated nearest-neighbor concepts. 
Yet, this mechanism fundamentally differs from our approach: ERC loss computes distances between static router weights to bound injected noise during resource-intensive pre-training. Conversely, kNN-MoE operates as a compute-efficient, inference-time adaptation. 
By retrieving the nearest neighbors of continuous input representations from an external memory, it directly reuses locally optimal expert assignments without the need for parameter updates.

\subsection{Non-Parametric and Retrieval-Augmented LMs}
Our work is closely related to non-parametric methods that augment neural networks with external data stores.
A prominent line of research involves kNN-LMs \citep{knnlm, knnlm_2}, which retrieve nearest-neighbor tokens from a data store of cached hidden states to interpolate the final output distribution.
Similarly, Retrieval-Augmented Generation models \citep{rag} retrieve textual chunks to condition the generation process.
Furthermore, \citet{cbdt_decoding} explored retrieving past high-utility decisions in reranking to guide output selection.

\section{Mixture-of-Experts}
\label{sec:moe}

Consider an MoE Transformer \citep{switch_transformer} model with $L$ layers, where a subset of indices $\mathcal{L}_{\text{MoE}} \subset \{1, \dots, L\}$ corresponds to MoE layers.
For any specific layer $\ell \in \mathcal{L}_{\text{MoE}}$, the module consists of $N$ expert networks $\{E^{(\ell)}_i\}_{i=1}^N$ and a router $R^{(\ell)}$.

Let $x^{(\ell)} \in \mathbb{R}^d$ denote the input hidden state of the MoE module at layer $\ell$, which is typically the output of the preceding attention block or normalization module.
The core mechanism of the MoE module employs a router to assign the input to the most relevant experts. 
The router is typically parameterized by a learnable weight matrix $W_r^{(\ell)} \in \mathbb{R}^{d \times N}$, and it generates the expert assignment (a sparse gating distribution) $a^{(\ell)}(x^{(\ell)}) \in \mathbb{R}^N$ via a TopK softmax function:
\begin{equation}
    \label{eq:router_gate}
    a^{(\ell)}(x^{(\ell)}) = \mathrm{TopK}\left(\mathrm{Softmax}(x^{(\ell)} W_r^{(\ell)})\right).
\end{equation}
The final output of the MoE module, denoted as $h^{(\ell)}$, is computed as the linear combination of the selected experts' outputs using the predicted expert assignment:
\begin{equation}
    \label{eq:moe_output}
    h^{(\ell)} = \sum_{i=1}^N a^{(\ell)}(x^{(\ell)})_i \cdot E^{(\ell)}_i(x^{(\ell)}).\nonumber
\end{equation}

During standard inference, the router weights $W_r^{(\ell)}$ remain fixed. This static routing strategy may limit adaptability when the test distribution diverges from the training distribution.

\section{Proposed Method: kNN-MoE}
\label{sec:knn_moe}

We propose \textbf{kNN-MoE} (Figure~\ref{fig:overview}), a retrieval-based framework that enhances routing decisions by leveraging a memory of historical router inputs and their locally optimal expert assignments.
We equip \emph{each} MoE module  $\ell \in \mathcal{L}_{\text{MoE}}$ with an independent memory store $\mathcal{M}^{(\ell)}$.
We omit the layer superscript $(\ell)$ as the method operates identically across layers.
kNN-MoE consists of two phases: (1) \textbf{Memory Construction} (\S\ref{sec:datastore}), which builds the expert assignment memory offline; and (2) \textbf{Confidence-Aware Adaptive Mixing} (\S\ref{sec:adaptive_gating}), which retrieves and utilizes stored expert assignments during inference.

\subsection{Memory Construction}
\label{sec:datastore}

The goal of memory construction is to create a key-value store $\mathcal{M} = \{(k_t, v_t)\}$ for the layer, where the key $k_t$ is a router input and the value $v_t$ is the locally optimal expert assignment that outperforms the original router.
This process involves two steps: data collection and deriving locally optimal expert assignments.

\paragraph{Data Collection}
We use a reference dataset $\mathcal{D}_{\text{ref}}$ consisting of sequences of ground-truth tokens.
Let $\mathbf{y} = (y_1, \dots, y_T)$ denote a sequence in $\mathcal{D}_{\text{ref}}$.
We run the frozen MoE model on these sequences in teacher-forcing mode.
For every time step $t \in [1,T]$, we collect the router input $x_t$ at the current layer and the corresponding target token $y_t$ from the sequence.
The router input serves as the retrieval key, i.e., $k_t = x_t$.

\paragraph{Deriving Locally Optimal Expert Assignments}
Instead of storing the model's original routing decision defined in Eq.~\ref{eq:router_gate}, we seek the locally optimal expert assignment that maximizes the prediction probability of the ground-truth token $y_t$.
We formulate this as an optimization problem for each token.

Let $\pi(r) = \mathrm{TopK}(\mathrm{Softmax}(r))$ denote the mapping from logits $r \in \mathbb{R}^N$ to sparse expert weights.
For a specific token $t$, we introduce a learnable logit vector $r_t$ to replace the parametric output $x_t W_r$.
We optimize $r_t$ to minimize the negative log-likelihood of the target token $y_t$:
\begin{align}
    \label{eq:gold_routing_obj}
    r_t^{*} &= \argmin_{r \in \mathbb{R}^N} \mathcal{L}_t(r), \\
    \mathcal{L}_t(r) &= - \log p_\theta\bigl(y_t \mid x_t, \text{routing}{=}\pi(r)\bigr), \nonumber
\end{align}
where $\theta$ represents the frozen parameters of the rest of the network.

We solve Eq.~\ref{eq:gold_routing_obj} via gradient descent.
We initialize $r_t$ with the original parametric logits, i.e., $r_t^{(0)} = x_t W_r$. We then perform $S$ steps of updates:
\begin{equation}
    r_t^{(s+1)} = r_t^{(s)} - \eta \nabla_r \mathcal{L}_t(r_t^{(s)}), ~~ \text{for } s = 0, \dots, S-1,\nonumber
\end{equation}
where $\eta$ is the learning rate.
Note that while we describe $r_t$ for a generic layer, this optimization is performed jointly across the entire model; specifically, the routing logits of all layers $\{r_t^{(\ell)}\}_{\ell \in \mathcal{L}_{\text{MoE}}}$ are updated simultaneously to minimize the loss.
Additionally, the optimization and data collection processes are conducted independently across different sequences in $\mathcal{D}_{\text{ref}}$.

The final optimized routing logits are $r_t^{(S)}$.
These logits serve as the target value in our memory, i.e., $v_t = r_t^{(S)}$.
They represent the ``(locally) optimal'' continuous routing signals that best predict the next token.

Finally, the memory for the current layer is constructed as:
\begin{equation}
    \mathcal{M} = \{(x_t, r_t^{(S)}) \mid t \in \mathcal{D}_{\text{ref}}\}.\nonumber
\end{equation}
This process is repeated for every layer in $\mathcal{L}_{\text{MoE}}$.

\subsection{Confidence-Aware Adaptive Mixing}
\label{sec:adaptive_gating}

During inference, for each MoE layer and each router input $x$,
we retrieve the set of $K$ nearest neighbors $\mathcal{N}(x) \subset \mathcal{M}$ from the corresponding memory constructed in \S\ref{sec:datastore}.

We formulate a non-parametric routing logit proposal, denoted as $r_{\text{mem}}(x)$, by aggregating the retrieved optimal logits according to their similarity to the input.
Specifically, let $\{(k_j, v_j)\}_{j=1}^K$ be the retrieved key-value pairs in $\mathcal{N}(x)$.
The memory-based logit proposal is computed as:
\begin{equation}
    \label{eq:mem_gate}
    r_{\text{mem}}(x) = \sum_{j=1}^K \frac{s(x, k_j)}{\sum_{m=1}^K s(x, k_m)} \, v_j,\nonumber
\end{equation}
where $s(\cdot, \cdot)$ is a similarity function.

To fuse the parametric router's original logits, $r(x) = x W_r^{(\ell)}$, with the non-parametric proposal $r_{\text{mem}}(x)$, we introduce an adaptive mixing coefficient $\lambda(x)$ that reflects retrieval confidence.
We quantify this confidence using the average similarity of the neighbors:
\begin{equation}
    \lambda(x) = \frac{1}{K} \sum_{j=1}^K s(x, k_j).\nonumber
\end{equation}

The final routing logits are obtained by linearly interpolating between the parametric and memory-based logits. We then apply the TopK-Softmax function $\pi(\cdot)$ to generate the final sparse expert assignment:
\begin{align}
    \label{eq:final_gating}
    r_{\text{final}}(x) &= (1 - \lambda(x)) r(x) + \lambda(x) r_{\text{mem}}(x),\\
    a_{\text{final}}(x) &= \pi(r_{\text{final}}(x)).\nonumber
\end{align}

Finally, we use $a_{\text{final}}(x)$ to compute the MoE layer output $h$:
\begin{equation}
    h = \sum_{i=1}^N a_{\text{final}}(x)_i \cdot E_i(x).\nonumber
\end{equation}
Performing the interpolation in the continuous logit space ensures that the final expert assignment strictly maintains the model's original Top-K sparsity constraint. Furthermore, this mixing ensures that when $\lambda(x) \approx 1$, the model trusts the retrieved routing signals to correct the router; conversely, when $\lambda(x) \approx 0$, it falls back to the parametric router to avoid retrieval noise.

\section{Experiments}
\label{sec:exp}

\subsection{Experimental Settings}
\label{sec:exp_setting}

\begin{table*}[!t]

    \centering

    \small{
    \begin{tabular}{llccccccc}

        \toprule

        \textbf{Model} & \textbf{Method} & \textbf{GPQA} & \textbf{MMLU} & \textbf{SuperGPQA} & \textbf{USMLE} & \textbf{MedMCQA} & \textbf{MBPP}& \textbf{Avg.} \\

        \midrule

        \multirow{6}{*}{\textbf{OLMoE}} 

        & Zero-shot & 27.27 & 46.77 & 13.02 & 32.81 & 35.57 & 30.00 & 26.82 \\

        & 5-shot & 21.72 & 33.06 & 11.09 & 31.31 & 25.36 &28.50&20.30\\

        & SFT & 21.72 & 46.89 & ~~\textbf{13.67}\textsuperscript{†} & ~~\textbf{37.84}\textsuperscript{†} & 34.78 & 31.60&~~27.26\textsuperscript{†}\\

        & SFT (Router Only) & 24.24 & 45.27 & 11.66 & 31.22 & 34.23 & 29.60&25.41\\

        & \textbf{kNN-MoE} & \textbf{29.80} & ~~\textbf{47.81}\textsuperscript{†} & ~~13.27\textsuperscript{†} & ~~35.04\textsuperscript{†} & ~~\textbf{37.01}\textsuperscript{†} &\textbf{32.20}& ~~\textbf{27.53}\textsuperscript{†}\\
        & \textbf{kNN-MoE (IVFPQ)} & \textbf{29.80} & ~~47.79\textsuperscript{†} & ~~13.14\textsuperscript{†} & ~~35.04\textsuperscript{†} & ~~36.98\textsuperscript{†} & 31.83&~~27.44\textsuperscript{†} \\

        \midrule

        \multirow{6}{*}{\textbf{GPT-OSS}} 

        & Zero-shot & 43.94 & 70.20 & 23.52 & 67.29 & 56.20 & 55.00& 43.27\\

        & 5-shot & 36.87 & 63.72 & 19.95 & 60.02 & 52.76&46.83 &38.61\\

        & SFT & 41.92 & 70.28 & ~~\textbf{24.83}\textsuperscript{†} & 65.61 & 56.06&48.83&~~43.85\textsuperscript{†} \\

        & SFT (Router Only) & 41.41 & 69.18 & 18.83 & 65.05 & 54.89 &45.00&40.09\\

        & \textbf{kNN-MoE} & \textbf{45.45} & 70.28 & ~~24.35\textsuperscript{†} & ~~\textbf{68.31}\textsuperscript{†} &~~57.06\textsuperscript{†} &\textbf{56.50}&~~\textbf{43.88}\textsuperscript{†}\\
        & \textbf{kNN-MoE (IVFPQ)} & 44.95 & \textbf{70.30} & ~~24.14\textsuperscript{†} & ~~68.13\textsuperscript{†} & ~~\textbf{57.14}\textsuperscript{†} & 56.17&~~43.77\textsuperscript{†}\\

        \midrule

        \multirow{6}{*}{\textbf{Qwen3}} 

        & Zero-shot & 41.41 & 78.59 & 34.85 & 75.30 & 62.51 & 59.80&52.93\\

        & 5-shot & 44.44 & ~~\textbf{80.48}\textsuperscript{†} & 35.56 & ~~77.17\textsuperscript{†} & ~~66.58\textsuperscript{†} &\textbf{62.33}&~~54.41\textsuperscript{†}\\

        & SFT & 39.90 & 79.08 & ~~\textbf{40.20}\textsuperscript{†} & ~~\textbf{80.80}\textsuperscript{†} & ~~66.60\textsuperscript{†}&55.33&~~\textbf{56.44}\textsuperscript{†} \\

        & SFT (Router Only) & 43.94 & 78.36 & 33.15 & 76.05 & ~~66.03\textsuperscript{†}&56.83&52.27\\

        & \textbf{kNN-MoE} & \textbf{44.95} & ~~78.86\textsuperscript{†} & 35.15 & ~~76.70\textsuperscript{†} & ~~66.65\textsuperscript{†} &61.83&~~53.66\textsuperscript{†}\\
        & \textbf{kNN-MoE (IVFPQ)} & 43.43 & 78.79 & 35.13 & 76.14 & ~~\textbf{66.70}\textsuperscript{†} & 61.83&~~53.61\textsuperscript{†}\\

        \bottomrule

    \end{tabular}}
    \caption{Performance comparison across three MoE models. † denotes a statistically significant improvement ($p<0.05$) over the Zero-shot baseline based on paired bootstrap resampling.}
    \label{tab:main_results}

\end{table*}

\paragraph{Datasets}
We evaluated kNN-MoE on general reasoning and domain-specific benchmarks: GPQA, SuperGPQA, MMLU, USMLE, MedMCQA, and MBPP. 
For each, we defined a disjoint test set $\mathcal{D}_{\text{test}}$ and a reference set $\mathcal{D}_{\text{ref}}$.
The same $\mathcal{D}_{\text{ref}}$ was reused across all methods:
kNN-MoE used it to construct the expert assignment memory, 5-shot \citep{icl} used it as the retrieval pool for in-context examples, and supervised fine-tuning (SFT) used it as the training dataset.
Table~\ref{tab:data_stats} detailed the splits.

\paragraph{Models}
We employed three representative MoE models: OLMoE (\code{allenai/OLMoE-1B-7B-0125-Instruct}) \citep{olmoe}, GPT-OSS (\code{openai/gpt-oss-20b}) \citep{gptoss}, and Qwen3 (\code{Qwen/Qwen3-30B-A3B-Instruct-2507}) \citep{qwen3}.
These models were selected because they incorporated modern MoE architectural designs and were computationally feasible to run within our research budget.

\paragraph{Baselines}
We compared kNN-MoE against two categories of baselines. 
First, we included \textbf{Zero-shot} and \textbf{5-shot} as standard, cost-effective inference baselines. 
Second, and importantly, we compared against \textbf{SFT}. 
Since kNN-MoE leveraged the reference set $\mathcal{D}_{\text{ref}}$ to improve performance, SFT served as the primary methodological baseline: similar to kNN-MoE, it utilized the reference set $\mathcal{D}_{\text{ref}}$ to optimize performance in an \emph{offline} stage (training), ensuring that the inference process remained efficient. 
This made SFT the most direct comparison for evaluating the efficacy of our offline memory construction.
\begin{itemize}
    \item \textbf{Zero-shot:} Zero-shot used the original MoE model with its parametric routers $R^{(\ell)}$ and a zero-shot QA prompt.
    \item \textbf{5-shot:} For each test instance in $\mathcal{D}_{\text{test}}$, we retrieved five examples from the reference set $\mathcal{D}_{\text{ref}}$ whose input questions were most similar to the test question. 
    Concretely, we encoded all questions using the \code{Qwen3-Embedding-0.6B} \citep{qwen3embedding} sentence embedding model and selected the five reference questions with the highest cosine similarity to the test question. 
    These examples, together with their correct answers, were concatenated before the test question to form the in-context prompt. 
    \item \textbf{SFT:} For SFT, we trained on $\mathcal{D}_{\text{ref}}$ using standard cross-entropy loss over the correct answers. 
    We applied LoRA adapters \citep{lora} to all linear layers.
    We set the maximum number of training epochs to $3$, split $\mathcal{D}_{\text{ref}}$ into an 85\% training split and a 15\% validation split, and selected the checkpoint with the lowest validation loss.
    During training and inference, SFT used a zero-shot QA prompt.

    \item \textbf{SFT (Router Only):} We fine-tuned only the router parameters $\{W_r^{(\ell)}\}_{\ell \in \mathcal{L}_{\text{MoE}}}$ on $\mathcal{D}_{\text{ref}}$, while freezing all expert and non-router parameters. 
    The training objective was the same supervised cross-entropy loss as in SFT.
    We used the same training protocol (including training epochs, data splitting, and checkpoint selection methods) and prompt as SFT.
\end{itemize}
See Appendix \ref{app:prompts} and \ref{app:training_details} for details on prompt templates and SFT training, respectively.

\paragraph{Note on Other Methods}
We excluded token-level retrieval methods like kNN-LM \citep{knnlm} from the main comparison due to the lack of efficient implementations compatible with modern LLM architectures.
Furthermore, we deferred the comparison with C3PO \citep{moe_opt3} to Appendix~\ref{sec:vs_c3po}. 
The rationale was that C3PO operated as a \emph{test-time adaptation} method, performing computationally intensive optimization \emph{during} inference for each input. 
In contrast, kNN-MoE aligned with the SFT paradigm by shifting the computational burden to an \emph{offline} memory construction, allowing for fast inference. Thus, we treated C3PO as a distinct category of method with different latency-accuracy trade-offs.

\paragraph{Implementation Details of kNN-MoE}
We used FAISS \citep{faiss} for retrieval and built one index per MoE layer, using the router inputs $\{x_t\}$ stored in that layer's memory $\mathcal{M}$ as keys.
The naive implementation of kNN-MoE relies on FP16 index storage and exact distance search. 
To balance search accuracy and efficiency, we consider an optimized version based on Product Quantization (PQ) \citep{pq}, Inverted File (IVF) clustering, and dimension reduction. 
Specifically, we quantize the FP16 index to 8-bit, search across only 32 centroids during retrieval, and compress the hidden dimension to one-eighth. 
We refer to this variant as \textbf{kNN-MoE (IVFPQ)}.
We used a single hyperparameter $K$ (neighbors per router) and shared it across layers; by default, we set $K=1$ based on validation (\S\ref{sec:k}).
We set the learning rate to $\eta = 2 \times 10^{-2}$ and used a single gradient descent step ($S=1$) for each token. 
We found that using more steps (e.g., $S=3$ or $S=10$) yielded very similar performance while incurring a substantially higher construction cost; we provided an ablation over $S$ in \S\ref{sec:step}.
The similarity function was an RBF kernel $s(x, k) = \exp(-\gamma \|x - k\|^2)$, with $\gamma$ set heuristically based on the average nearest neighbor distance in the memory. 

\paragraph{Evaluation}
We reported \textbf{accuracy} (\%) on $\mathcal{D}_{\text{test}}$ for all datasets, using the official benchmark answers.
To ensure a fair comparison, we kept the answer parsing identical across all methods. 
Additionally, we reported the overall accuracy (\textbf{Avg.}), which is calculated over a concatenated dataset comprising all individual test benchmarks. 
To assess statistical significance, we performed paired bootstrap resampling \citep{pbs} comparing all methods against the zero-shot baseline. 

\subsection{Results}

\label{sec:exp_results}


\paragraph{Comparison with Zero-shot and SFT (Router Only)} 
Across all models and datasets, kNN-MoE consistently improves over both Zero-shot and SFT (Router Only). 
These trends indicate that refining expert assignment via retrieved cases is more effective than solely relying on the fixed parametric router. 
It is worth noting that the accuracy of the IVFPQ variant closely matches that of the naive implementation, experiencing only a slight degradation. 
Importantly, it still consistently outperforms both Zero-shot and SFT (Router Only).

\paragraph{Comparison with 5-shot}
Compared to 5-shot, kNN-MoE exhibits more stable behavior across models. 
On Qwen3~30B, both 5-shot and kNN-MoE can improve over Zero-shot;
but for GPT-OSS and OLMoE, 5-shot often degrades performance relative to Zero-shot. 
A likely reason is that concatenating the five nearest reference examples substantially lengthens the input sequence, pushing these models into length regimes where their performance is less robust and amplifying the effect of retrieval noise. 

\paragraph{Comparison with SFT}
While kNN-MoE outperforms SFT in certain scenarios, its relative performance varies. 
In settings with limited reference sets (GPQA, USMLE and MBPP), SFT occasionally underperforms compared to Zero-shot. 
Our analysis in Appendix~\ref{sec:lb} indicates that this degradation cannot be attributed to a collapse in routing load balancing; therefore, we attribute it to potential overfitting induced by the small dataset size. 
In contrast, our kNN-MoE consistently outperforms Zero-shot in these data-scarce settings.

\paragraph{Statistical Significance}
Based on our paired bootstrap resampling tests, although improvements on certain individual benchmarks are not always statistically significant, kNN-MoE demonstrates a statistically significant improvement on the overall concatenated dataset (\textbf{Avg.}). 
This overall significance across diverse tasks solidifies our main conclusion. 
Furthermore, on benchmarks where kNN-MoE's gains are not statistically significant (e.g., GPQA), competitive methods such as SFT similarly fail to achieve significant improvements. This suggests that the lack of significance on these specific tasks is likely not a limitation of our routing refinement framework, but rather indicates that existing models struggle to extract further performance gains relying solely on the information provided in the given reference sets.


\section{Discussion}

\label{sec:analysis}

\subsection{Why Case-Based MoE Works}

\label{sec:why}

Our main hypothesis is that retrieval is most beneficial when the parametric router is uncertain, which is often correlated with increased perplexity.
To test this, we bucketed test instances into three equal-sized groups based on the perplexity (PPL) of the original model (with zero-shot prompt) (High / Mid / Low PPL), and reported the accuracy gain of kNN-MoE over Zero-shot within each bucket.
Table~\ref{tab:why_works} reports results on OLMoE.

\begin{table}[h]
\setlength\tabcolsep{3pt}
    \centering
    \small{
    \begin{tabular}{lccc}
        \toprule
        \textbf{Benchmark} & High PPL & Mid PPL & Low PPL  \\
        \midrule
        GPQA       & +6.15 & +2.99 & -1.52 \\
        SuperGPQA  & +0.17 & +0.65 & -0.06 \\
        MMLU       & +2.07 & +1.15 & -0.13 \\
        USMLE      & +6.32 & +1.41 & -1.13\\
        MedMCQA    & +6.04 & +1.13 & +0.56\\
        \bottomrule
    \end{tabular}}
    \caption{Accuracy gains (kNN-MoE minus Zero-shot) bucketed by perplexity (PPL) of the original model (Zero-shot), using OLMoE.}
    \label{tab:why_works}
\end{table}

kNN-MoE usually yields the largest gains on High-PPL instances across all benchmarks (e.g., +6.15 on GPQA and +6.32 on USMLE), while the improvements on Mid-PPL instances are smaller but still consistently positive.
In contrast, the gains on Low-PPL instances are close to zero and can become slightly negative on some datasets.
The slight degradation on a subset of Low-PPL instances suggests that kNN-MoE can occasionally introduce noise when the router prediction is strong.

Overall, we hypothesize that the strong correlation between perplexity and kNN-MoE gains stems from the fact that higher perplexity indicates a greater deviation of the input from the router's data distribution. 
Consequently, the expert assignment predicted by the router becomes unreliable, whereas kNN-MoE improves the assignment.

\begin{table}[h]
    \centering
    \small{
    \begin{tabular}{lccc}
        \toprule
        Method & GPQA & MMLU & USMLE  \\
        \midrule
        kNN-MoE &29.80& 47.81& 35.04\\
        kNN-MoE (Selective) &30.30& 47.85& 35.41\\
        \bottomrule
    \end{tabular}}
    \caption{Performance comparison on OLMoE between standard kNN-MoE and a \textbf{Selective} variant. The Selective variant falls back to the zero-shot baseline for inputs in the lowest perplexity tercile.}
    \label{tab:why_works2}
\end{table}

Inspired by this observation, we investigated a variant termed \textbf{kNN-MoE (Selective)}.
This variant dynamically deactivates retrieval for inputs where Zero-shot is confident (specifically, the bottom 33\% of test instances by perplexity) and falls back to the parametric router.
As shown in Table~\ref{tab:why_works2}, this strategy yields marginal improvements.
However, it introduces significant computational overhead, as it requires a preliminary forward pass to estimate baseline perplexity.
Given the negligible performance gain relative to the added cost, we retained the standard kNN-MoE as our primary approach.

\subsection{Impact of Reference Set Size $|\mathcal{D}_{\text{ref}}|$}
\label{sec:data_size}
\begin{table}[h]
    \centering
    \small{
    \begin{tabular}{lccc}
        \toprule
        $|\mathcal{D}_{\text{ref}}|$ & OLMoE & GPT-OSS & Qwen3  \\
        \midrule
        $0$ & 35.57 & 56.20 & 62.51   \\
        $0.25$k & 35.99 & 56.41 & 66.01 \\
        $0.50$k & 36.87 & 56.92 & 66.68 \\
        $1.00$k & 37.01 & 57.06 & 66.65 \\
        \bottomrule
    \end{tabular}}
    \caption{Accuracy (\%) of kNN-MoE on MedMCQA as a function of the reference set size $|\mathcal{D}_{\text{ref}}|$ used for memory construction. $|\mathcal{D}_{\text{ref}}|=0$ corresponds to Zero-shot (no memory).}
    \label{tab:size}
\end{table}

We next varied the size of the reference set used to build the per-layer memory and evaluate kNN-MoE on MedMCQA.
Table~\ref{tab:size} shows that larger memories generally improve performance for all three models.
When $|\mathcal{D}_{\text{ref}}|=0$, kNN-MoE degenerates to the zero-shot baseline.
As $|\mathcal{D}_{\text{ref}}|$ increases, accuracy improves monotonically, with diminishing returns beyond roughly $0.5$k--$1$k examples.

\subsection{Impact of the Number of Neighbors ($K$)}

\label{sec:k}

\begin{table}[h]
    \centering
    \small{
    \begin{tabular}{lccc}
        \toprule
        $K$ & OLMoE & GPT-OSS & Qwen3  \\
        \midrule
        $1$ & 37.01 & 57.06 & 66.65   \\
        $3$ & 36.89 & 56.55 & 65.76 \\
        $5$ & 36.02 & 56.74 & 65.98 \\
        $10$ & 36.34 & 56.11 & 65.19 \\
        \bottomrule
    \end{tabular}}
    \caption{Accuracy (\%) of kNN-MoE on MedMCQA using different numbers of neighbors $K$.}
    \label{tab:k}
\end{table}

We investigated the sensitivity of kNN-MoE to the value of $K$.
Table~\ref{tab:k} presents the accuracy on MedMCQA for varying $K \in \{1, 3, 5, 10\}$.
Contrary to typical non-parametric approaches \citep{knnlm, cbdt_decoding} where aggregating multiple neighbors (e.g., $K \ge 5$) helps reduce variance and smooth out noise, we observe that performance consistently peaks at $K=1$ across all three models.
As $K$ increases, accuracy generally degrades.

We interpret this essentially as a negative result regarding the benefit of neighbor aggregation in the context of MoE routing.
This phenomenon suggests that the locally optimal expert assignments responsible for positive outcomes are rare within the memory $\mathcal{M}$.
In other words, for a given query, there is often only a single (or very few) historical case that provides a truly beneficial routing signal.
Consequently, forcing the retrieval of additional neighbors ($K > 1$) likely incorporates cases with lower relevance or conflicting expert assignments.
Instead of reinforcing the correct decision, these additional neighbors dilute the strong signal provided by the nearest neighbor, thereby introducing noise that harms the final model performance.
Based on this finding, we adopt $K=1$ as the default setting for all other experiments.

\subsection{Impact of Gradient Descent Steps ($S$)}
\label{sec:step}
\begin{table}[h]
    \centering
    \small{
    \begin{tabular}{lccc}
        \toprule
        $S$ & OLMoE & GPT-OSS & Qwen3  \\
        \midrule
        $1$ & 37.01 & 57.06 & 66.65   \\
        $3$ & 37.05 & 57.12 & 66.58 \\
        $10$ & 36.98 & 57.01 & 66.71 \\
        \bottomrule
    \end{tabular}}
    \caption{Accuracy (\%) of kNN-MoE on MedMCQA with varying gradient descent steps $S$ used for memory construction.}
    \label{tab:s}
\end{table}

We studied the impact of the number of gradient descent steps $S$ used during the \textit{Memory Construction} phase (\S\ref{sec:datastore}) to derive the locally optimal expert assignment.
Intuitively, more optimization steps should lead to a lower negative log-likelihood on the reference tokens, potentially yielding higher-quality targets for the memory.
We compared the performance of kNN-MoE constructed with $S \in \{1, 3, 10\}$ steps.

Table~\ref{tab:s} summarizes the results on MedMCQA.
We observe that increasing $S$ beyond a single step yields negligible improvements in downstream accuracy.
We hypothesize that the initial gradient direction $\nabla_r \mathcal{L}_t(r_t^{(0)})$ already captures the most critical information regarding which experts are under-utilized or should be promoted.
Since the memory stores discrete locally optimal assignment values $v_t$, slight refinements to the logits $r_t$ in subsequent steps ($S>1$) may not significantly alter the top-k ranking or the resulting routing distribution used as targets.
Crucially, the computational cost of memory construction scales linearly with $S$.
Setting $S=1$ allows for rapid memory construction (as shown in \S\ref{sec:cost}) without compromising model performance.
Therefore, we adopted $S=1$ as the default setting for efficiency.

\subsection{Memory Construction and Inference Costs}
\label{sec:cost}
We investigated the memory construction and inference costs of kNN-MoE.
Table~\ref{tab:mem_time} compares the preparation time on OLMoE with $|\mathcal{D}_{\text{ref}}|=1.00$k, all conducted on a single NVIDIA A100 80GB.
Constructing the kNN-MoE memory took 0.46 hours, which was substantially faster than both SFT and SFT (Router Only).
This speedup is mainly driven by optimizing token-specific routing logits (rather than updating model parameters) and using only one gradient descent step per token.
Additionally, the IVFPQ variant introduces an extra $0.21$h for index quantization and compression. 
However, this minor overhead does not alter our overall conclusion regarding the significant preparation time advantage over SFT.

\begin{table}[h]
    \centering
    \small{
    \begin{tabular}{lccc}
        \toprule
        $|\mathcal{D}_{\text{ref}}|$ & SFT & \small{SFT (Router Only)} & kNN-MoE  \\
        \midrule
        $1.00$k & 1.34h & 1.33h & 0.46h \\
        \bottomrule
    \end{tabular}}
    \caption{Memory construction time on OLMoE with $|\mathcal{D}_{\text{ref}}|=1$k. Other reference sizes scale approximately linearly with $|\mathcal{D}_{\text{ref}}|$.}
    \label{tab:mem_time}
\end{table}

At inference time, kNN-MoE incurs additional latency due to per-token nearest neighbor retrieval.
Table~\ref{tab:infer_speed} reports the average latency per example as a function of $|\mathcal{D}_{\text{ref}}|$.
As expected, latency increases with $|\mathcal{D}_{\text{ref}}|$ because each router query searches a larger index.
With $|\mathcal{D}_{\text{ref}}|=1.00$k and $K=1$, the overhead of naive kNN-MoE remains moderate.
Fortunately, our IVFPQ variant successfully mitigates this inference time overhead. Even with $|\mathcal{D}_{\text{ref}}|=1.00$k, it only increases the inference latency by approximately 3.67\% relative to the zero-shot baseline.

\begin{table}[h]
    \centering
    \small{
    \begin{tabular}{lcc}
        \toprule
        $|\mathcal{D}_{\text{ref}}|$ & kNN-MoE & kNN-MoE (IVFPQ)  \\
        \midrule
        $0$ &  5.99s  & 5.99s \\
        $0.25$k &  6.19s & 6.14s \\
        $0.50$k &  6.41s & 6.16s\\
        $1.00$k &  6.83s & 6.21s\\
        \bottomrule
    \end{tabular}}
    \caption{Inference speed (seconds per example) using different reference set sizes $|\mathcal{D}_{\text{ref}}|$.}
    \label{tab:infer_speed}
\end{table}

Table~\ref{tab:vram} illustrates the VRAM consumption during inference. While the VRAM usage of the naive kNN-MoE implementation increases significantly as $|\mathcal{D}_{\text{ref}}|$ grows, the IVFPQ variant exhibits a substantially slower growth rate. This efficiency is achieved through vector quantization, dimension reduction, and clustering. Even at $|\mathcal{D}_{\text{ref}}|=1.00$k, the IVFPQ variant only increases VRAM usage by approximately 9.43\% compared to the baseline.
\begin{table}[h]
    \centering
    \small{
    \begin{tabular}{lcc}
        \toprule
        $|\mathcal{D}_{\text{ref}}|$ & kNN-MoE & kNN-MoE (IVFPQ)  \\
        \midrule
        $0$ &  16.02GB  & 16.02GB \\
        $0.25$k &  19.31GB & 16.21GB \\
        $0.50$k &  22.47GB & 16.94GB\\
        $1.00$k &  28.81GB & 17.53GB\\
        \bottomrule
    \end{tabular}}
    \caption{VRAM consumption (GB) during inference across different reference set sizes $|\mathcal{D}_{\text{ref}}|$.}
    \label{tab:vram}
\end{table}

\section{Conclusions and Future Work}
\label{sec:conclusions}
We introduced kNN-MoE, a framework that augments MoE parametric routers with a non-parametric memory of locally optimal expert assignments derived from a labeled reference set. By interpolating router predictions with retrieved nearest-neighbor assignments using a similarity-based coefficient, kNN-MoE improves routing without parameter updates. 
Across three models and six benchmarks, it outperforms Zero-shot and router-only SFT, achieving SFT-competitive results with substantially less preparation time.

Future work will focus on two directions. First, we aim to accelerate retrieval via memory pruning to reduce computational overhead while preserving accuracy. Second, we plan to extend kNN-MoE to unlabeled reference sets by utilizing \emph{LLM-as-a-judge} \citep{llm-as-judge} to generate pseudo-labels for constructing locally optimal expert assignments.

\section*{Limitations}
The primary limitation of our current framework lies in its dependence on a labeled reference set ($\mathcal{D}_{\text{ref}}$) that shares a similar distribution with the target test data.
Our experimental success is predicated on the assumption that such accessible, high-quality data is available.
However, in practical, open-ended deployment scenarios, acquiring domain-specific labeled data can be challenging or expensive.
Consequently, this dependency may restrict the applicability of kNN-MoE in strictly zero-resource settings or when facing severe distribution shifts where no relevant historical cases exist.
As mentioned in Section~\ref{sec:conclusions}, leveraging \emph{LLM-as-a-judge} to generate pseudo-labels offers a potential pathway to mitigate this constraint by utilizing unlabeled data; however, we leave the validation of this semi-supervised approach to future work.

Additionally, while kNN-MoE avoids the high cost of parameter updates, the non-parametric retrieval mechanism introduces necessary overheads.
Specifically, the inference latency increases due to neighbor search, and the memory storage requirements scale with the size of the reference set.
While currently manageable for the scales investigated in this paper, these factors may necessitate further optimization, such as index pruning, for deployment in strictly resource-constrained environments.

Finally, our experiments involved only a limited number of datasets and models, which means our conclusions are currently restricted to the relevant tasks and MoE models with up to 128 experts.

\bibliography{custom}

\clearpage

\appendix

\section{Prompt Templates}
\label{app:prompts}

To ensure reproducibility, we provide the exact prompt templates used for our Zero-shot and 5-shot inference baselines.
Please note that the templates below illustrate the case for multiple-choice questions with four options (A--D). For questions with a different number of options, the list of choices is adjusted accordingly while maintaining the same overall format.

\subsection{Zero-shot Prompt}
For all models and test sets in the Zero-shot setting, we formatted the input question and options using the template below. 
The placeholders \texttt{\{question\}} and \texttt{\{choices\}} are substituted with the specific instance data.

\begin{figure}[h]
\centering
\begin{lstlisting}[
  basicstyle=\ttfamily\small\selectfont,
  breaklines=true,
  breakatwhitespace=true,
  aboveskip=0pt,
  belowskip=0pt,
  backgroundcolor=\color{gray!10},
  frame=single,
  framerule=0.4pt,
  framesep=4pt,
  rulecolor=\color{gray!40}
]
What is the correct answer to this question: {question}

Choices:
(A) {choices[0]}
(B) {choices[1]}
(C) {choices[2]}
(D) {choices[3]}

Answer with the format: The correct answer is (X).
\end{lstlisting}
\caption{Prompt template for the zero-shot setting.}
\label{fig:prompt-for-zeroshot-prompting}
\end{figure}

\subsection{5-shot Prompt}
For the 5-shot setting, we constructed the prompt by concatenating five retrieved expert examples followed by the target question. 
Similar to the zero-shot case, the example below assumes four choices per question.

\begin{figure}[h]
\centering
\begin{lstlisting}[
  basicstyle=\ttfamily\small\selectfont,
  breaklines=true,
  breakatwhitespace=true,
  aboveskip=0pt,
  belowskip=0pt,
  backgroundcolor=\color{gray!10},
  frame=single,
  framerule=0.4pt,
  framesep=4pt,
  rulecolor=\color{gray!40}
]
Here are some example expert multiple-choice questions. After the examples, answer the final question.

Question: {s1.question}
Choices: 
(A) {s1.choices[0]}
(B) {s1.choices[1]}
(C) {s1.choices[2]}
(D) {s1.choices[3]}
The correct answer is ({s1.answer})

... [3 more examples] ...

Question: {s5.question}
Choices:
(A) {s5.choices[0]}
(B) {s5.choices[1]}
(C) {s5.choices[2]}
(D) {s5.choices[3]}
The correct answer is ({s5.answer})

Question: {target.question}
Choices: 
(A) {target.choices[0]}
(B) {target.choices[1]}
(C) {target.choices[2]}
(D) {target.choices[3]}
\end{lstlisting}
\caption{Prompt template for the 5-shot setting.}
\label{fig:prompt-for-5shot-prompting}
\end{figure}

\section{Training Implementation Details}
\label{app:training_details}

Our supervised fine-tuning baselines---SFT and SFT (Router Only)---were implemented based on the TRL library\footnote{\url{https://github.com/huggingface/trl}}.
In the SFT (Router Only) setting, we constrained the optimizer to update only the router parameters.

Table~\ref{tab:hyperparams} summarizes the hyperparameters used for training.

\section{Additional Discussion}
\subsection{Comparison with C3PO}
\label{sec:vs_c3po}

\begin{table}[h]
    \centering
    \small{
    \begin{tabular}{lccc}
        \toprule
        Method & GPQA & MMLU & USMLE  \\
        \midrule
        Zero-shot &27.27& 46.77& 32.81\\
        C3PO &24.24&\textbf{48.04}& 32.93\\
        kNN-MoE &\textbf{29.80}& 47.81& \textbf{35.04}\\
        \bottomrule
    \end{tabular}}
    \caption{Performance comparison with the state-of-the-art method C3PO using the OLMoE model. kNN-MoE demonstrates superior sample efficiency on smaller datasets (GPQA, USMLE).}
    \label{tab:vs_c3po}
\end{table}

We compared kNN-MoE with C3PO \citep{moe_opt3}, a state-of-the-art test-time routing adaptation method.
Table~\ref{tab:vs_c3po} presents the accuracy on OLMoE across three benchmarks with varying reference set sizes.
We observe distinct behaviors depending on the availability of reference data.
On benchmarks with limited reference sets, (GPQA and USMLE), C3PO yields marginal gains or even performance degradation compared to the zero-shot baseline.
In contrast, kNN-MoE achieves consistent improvements in these data-scarce regimes.
However, on MMLU, which provides a larger reference set ($|\mathcal{D}_{\text{ref}}| \approx 1.81$k), C3PO outperforms kNN-MoE.

We attribute these findings to the difference in sample efficiency.
C3PO relies on a filtering mechanism that only utilizes reference examples correctly predicted by the baseline model to guide the router update.
This significantly reduces the effective training signal when the reference set is small or the task is difficult.
Conversely, kNN-MoE exploits \emph{all} retrieved neighbors by computing the locally optimal expert assignment that maximizes the likelihood of the ground-truth token, regardless of the original model's correctness.
This capability allows kNN-MoE to extract richer supervision signals from small datasets, making it more robust for specialized domains with limited data.

\subsection{Load Balancing Analysis}
\label{sec:lb}

We analyze the changes in the average load balancing loss across the entire test set for different methods to investigate the reasons behind the performance degradation of SFT. We report the results evaluated on GPQA using the OLMoE model. 

As shown in Table~\ref{tab:lb}, the load balancing losses for Zero-shot, SFT, SFT (Router Only), and kNN-MoE are highly similar. This indicates that the routing load balancing in SFT and its Router-Only variant did not collapse. This is an expected outcome, as the load balancing loss was incorporated as a constraint into the overall training objective. Therefore, we conjecture that the performance degradation observed in SFT and SFT (Router Only) is primarily due to overfitting caused by the limited size of the training dataset.

\begin{table}[h]
    \centering
    \small{
    \begin{tabular}{cccc}
        \toprule
         Zero-shot & SFT & \small{SFT (Router Only)} & kNN-MoE  \\
        \midrule
         8.27 & 8.29 & 8.28 & 8.29 \\
        \bottomrule
    \end{tabular}}
    \caption{Average load balancing loss on the GPQA test set using the OLMoE model across different methods.}
    \label{tab:lb}
\end{table}

\subsection{Impact of Choice of $s(\cdot, \cdot)$}
\label{sec:s_func}
\begin{table}[h]
    \centering
    \small{
    \begin{tabular}{lccc}
        \toprule
        $s(\cdot, \cdot)$ & OLMoE & GPT-OSS & Qwen3  \\
        \midrule
        RBF & \textbf{37.01} & \textbf{57.06} & 66.65   \\
        RBF + L2 Norm & 36.52 & 56.19 & \textbf{66.68} \\
        Cosine & 36.69 & 56.01 & 66.65 \\
        \bottomrule
    \end{tabular}}
    \caption{Accuracy (\%) of kNN-MoE on MedMCQA using different similarity functions $s(\cdot, \cdot)$. ``RBF + L2 Norm'' applies the RBF kernel to L2-normalized router inputs, isolating the contribution of vector magnitude.}
    \label{tab:s_func_results}
\end{table}
We analyzed the sensitivity of kNN-MoE to the choice of the similarity function $s(x, x')$.
Specifically, we compared our default RBF kernel, which relies on Euclidean distance, against cosine similarity, which measures the cosine of the angle between vectors.
Table~\ref{tab:s_func_results} presents the results on MedMCQA.
We observe that the RBF kernel outperforms or matches cosine similarity.
 
We conjecture that this disparity stems from the way each metric handles vector magnitude.
Cosine similarity normalizes the input vectors, effectively projecting them onto a unit hypersphere and discarding the information encoded in their norms: $s_{\text{cosine}}(x, k) = \frac{x \cdot k}{\|x\| \|k\|}$.
In contrast, the RBF kernel depends on the distance $\|x - k\|^2$, which inherently preserves information about the scale (magnitude) of the router inputs.
The superior performance of RBF on OLMoE and GPT-OSS suggests that for these architectures, the magnitude of the router input vector $x$ likely exhibits significant variance and carries discriminative signals for expert assignment---information that is discarded when using cosine similarity.
Conversely, the insensitivity of Qwen3 to the results with different metrics imply that its router inputs may be more uniform in magnitude or that its routing policy is primarily direction-dependent.
 
To directly test this hypothesis, we additionally evaluated an \textbf{RBF + L2 Norm} variant, in which the RBF kernel is applied to L2-normalized router inputs.
This variant preserves the kernel form of the original RBF but removes magnitude information, isolating it as the only difference from standard RBF.
As shown in Table~\ref{tab:s_func_results}, once magnitude is removed, the performance of RBF on OLMoE and GPT-OSS drops to essentially the same level as cosine similarity, while Qwen3 remains nearly unchanged.
This result directly confirms that the gap between RBF and cosine on OLMoE and GPT-OSS is attributable to the magnitude of the router inputs, and that for Qwen3 the router inputs are effectively magnitude-uniform---so removing magnitude has no effect.
We therefore retain the standard RBF kernel as the default in kNN-MoE, since it captures magnitude information that is useful on at least a subset of models while not hurting magnitude-uniform models such as Qwen3.

\subsection{Further Analysis on the Number of Neighbors ($K$)}
\label{sec:k_analysis}
 
As shown in \S\ref{sec:k}, kNN-MoE empirically prefers $K=1$ across all three models, which appears to conflict with the intuition behind kNN aggregation.
To better understand this phenomenon, we conducted an additional analysis quantifying how much of the aggregated similarity mass is concentrated on the single nearest neighbor under a $K=5$ retrieval setting.
Specifically, for each retrieved set $\{(k_j, v_j)\}_{j=1}^{5}$, we computed the share of total similarity captured by the top-1 neighbor:
\begin{equation}
    \text{Top-1 Share} = \frac{s(x, k_1)}{\sum_{j=1}^{5} s(x, k_j)}, \nonumber
\end{equation}
where $k_1$ denotes the most similar neighbor.
Table~\ref{tab:top1_share} reports the average Top-1 Share over the corresponding test set.
 
\begin{table}[h]
    \centering
    \small{
    \begin{tabular}{lccc}
        \toprule
        Dataset & OLMoE & GPT-OSS & Qwen3  \\
        \midrule
        GPQA     & 0.91 & 0.82 & 0.94 \\
        MMLU     & 0.78 & 0.76 & 0.83 \\
        MedMCQA  & 0.88 & 0.85 & 0.93 \\
        \bottomrule
    \end{tabular}}
    \caption{Average share of total similarity mass held by the top-1 neighbor under $K=5$ retrieval. Across all settings, the nearest neighbor already accounts for the overwhelming majority of the aggregated similarity.}
    \label{tab:top1_share}
\end{table}
 
Across all model--benchmark pairs, the top-1 neighbor accounts for $76\%$--$94\%$ of the total similarity mass.
This indicates that even when $K>1$, the aggregated memory-based proposal $r_{\text{mem}}(x)$ is already dominated by the nearest neighbor, while the marginal contribution of farther neighbors is small in magnitude.
However, as Table~\ref{tab:k} shows, increasing $K$ consistently degrades accuracy.
This suggests that the additional neighbors are small in magnitude but harmful in direction: they inject conflicting routing signals rather than reinforce the correct one, which is consistent with our earlier interpretation that beneficial expert assignments are sparse in the memory.
 
We further emphasize that the empirical preference for $K=1$ is a property of the current models and reference sets rather than a structural property of the framework.
We deliberately retain the general $K$-neighbor formulation in \S\ref{sec:adaptive_gating} so that the method remains applicable to future settings (e.g., larger or more diverse memories, alternative aggregation schemes, or domains in which beneficial routing signals are less sparse), even though the empirical sweet spot in our current setup happens to be $K=1$.

\section{Impact of Random Seeds}
 
We analyze the impact of random seeds on the performance of different methods. We restrict our analysis to SFT and its variants, as these are the only methods involving training where random LoRA parameter initialization can introduce performance variance. Other methods (e.g., Zero-shot and kNN-MoE) are deterministic in this regard. Due to computational constraints, we report the mean and standard deviation of the accuracy on a subset of benchmarks using the OLMoE model.
 
As shown in Table~\ref{tab:seed}, SFT and its variants exhibit a certain degree of accuracy fluctuation across different runs. Nevertheless, this variance is relatively small and does not alter our primary conclusions.
 
\begin{table}[h]
    \centering
    \small{
    \begin{tabular}{ccc}
        \toprule
         \textbf{Benchmark} & SFT & \small{SFT (Router Only)}  \\
        \midrule
         GPQA & 22.56$\pm$1.89 & 23.39$\pm$1.89 \\
         MMLU & 46.21$\pm$1.02 & 45.67$\pm$0.95  \\
         USMLE & 38.29$\pm$0.51 & 31.59$\pm$0.32  \\
        \bottomrule
    \end{tabular}}
    \caption{Accuracy (mean $\pm$ standard deviation, \%) across different random seeds for SFT and SFT (Router Only) using the OLMoE model.}
    \label{tab:seed}
\end{table}

\begin{table*}[h]
    \centering
    \begin{tabular}{lc}
        \toprule
        \textbf{Configuration} & \textbf{Value} \\
        \midrule
        \multicolumn{2}{c}{\textit{Optimization}} \\
        Optimizer & AdamW \\
        Batch Size & 16 \\
        Weight Decay & 0.01 \\
        Learning Rate Schedule & Linear Decay \\
        \midrule
        \multicolumn{2}{c}{\textit{LoRA Parameters}} \\
        Rank ($r$) & 32 \\
        Alpha ($\alpha$) & 32 \\
        Dropout & 0.0 \\
        Bias & None \\
        \midrule
        \multicolumn{2}{c}{\textit{Hardware}} \\
        GPU & 1 $\times$ NVIDIA A100 (80GB) \\
        \bottomrule
    \end{tabular}
    \caption{Hyperparameters and configuration for SFT and SFT (Router Only) experiments.}    \label{tab:hyperparams}
\end{table*}

\begin{table*}[t]
    \centering
    \begin{tabular}{lcccc}
        \toprule
        \textbf{Benchmark} & \textbf{Test Source} & \textbf{Ref. Source} & $|\mathcal{D}_{\text{test}}|$ & $|\mathcal{D}_{\text{ref}}|$ \\
        \midrule
        GPQA       & GPQA Diamond & GPQA Main (filtered) & $0.20$k & $0.20$k \\
        MMLU       & Test subset & Train subset & $14.00$k & $1.81$k \\
        SuperGPQA  & Held-out split & Random subset & $23.50$k & $3.00$k \\
        USMLE      & Test subset & Train subset & $1.27$k & $0.20$k \\
        MedMCQA    & Test subset & Train subset & $6.15$k & $1.00$k \\
        MBPP    & Held-out split & Random subset & $0.60$k & $0.37$k \\
        \bottomrule
    \end{tabular}
    \caption{Statistics of evaluation benchmarks. $\mathcal{D}_{\text{test}}$ was used for reporting accuracy; $\mathcal{D}_{\text{ref}}$ served as the shared reference split across methods (for memory construction in kNN-MoE, retrieval in 5-shot, and SFT).}
    \label{tab:data_stats}
\end{table*}
 
\end{document}